\pdfoutput=1

\documentclass[10pt,journal,twocolumn]{IEEEtran}
\ifCLASSOPTIONcompsoc
  \usepackage[nocompress]{cite}
\else
  \usepackage{cite}
\fi
\ifCLASSINFOpdf
\else
\fi

\usepackage{times}
\usepackage{epsfig}
\usepackage{graphicx}
\usepackage{amsmath}
\usepackage{amssymb}
\usepackage{eucal,bibspacing}
\usepackage[bf,scriptsize]{caption}
\usepackage{cite}
\usepackage{cs}
\usepackage{mathsymb}
\usepackage{textcomp}
\usepackage{verbatim}
\usepackage{subfigure}
\usepackage{url}
\usepackage{graphicx}
\usepackage{multirow}
\usepackage[super]{nth}
\usepackage{colortbl}
\graphicspath{{./}{./Fig/}{./Figs/}{./Fig/eps/}{./Fig/pdf/}}

\usepackage{algorithm2e}

\let\savedalgorithm\algorithm
\let\savedendalgorithm\endalgorithm

\usepackage{algorithm}

\def\bw{{\boldsymbol w}}

\begin{document}

\title{PersonNet: Person Re-identification with  Deep Convolutional Neural Networks}

\author{Lin Wu, Chunhua Shen, Anton van den Hengel%
\IEEEcompsocitemizethanks{\IEEEcompsocthanksitem Authors are with The University of
Adelaide, Australia; and Australian Research Council Centre of Excellence
for Robotic Vision;
Corresponding author: Chunhua Shen (chunhua.shen@adelaide.edu.au)\protect\\
}%
}

\IEEEtitleabstractindextext{%
\begin{abstract}
In this paper, we propose a  deep end-to-end neural network to simultaneously learn high-level features and
a corresponding similarity metric for person re-identification.
The network takes a pair of raw RGB images as input, and outputs a similarity value indicating whether
the two input images depict the same person.
A layer of computing neighborhood range differences across two input images is employed to capture local relationship between
patches \cite{JointRe-id}. This operation is to seek a robust feature from input images. By increasing the depth to 10 weight layers and using very small (3$\times$3) convolution filters, our architecture achieves a remarkable improvement on the prior-art configurations. Meanwhile, an adaptive Root-Mean-Square (RMSProp) gradient decent algorithm is integrated into our architecture, which is beneficial to deep nets. Our method consistently outperforms state-of-the-art on two large datasets (CUHK03 and Market-1501), and a medium-sized data set (CUHK01).
\end{abstract}

\begin{IEEEkeywords}
Person re-identification, Convolutional neural networks, Deep metric learning.
\end{IEEEkeywords}}

\maketitle

\IEEEdisplaynontitleabstractindextext

\IEEEpeerreviewmaketitle

\section{Introduction}\label{sec:introduction}

\begin{figure}[t]
    \centering
        \includegraphics[width=3in,height=2in]{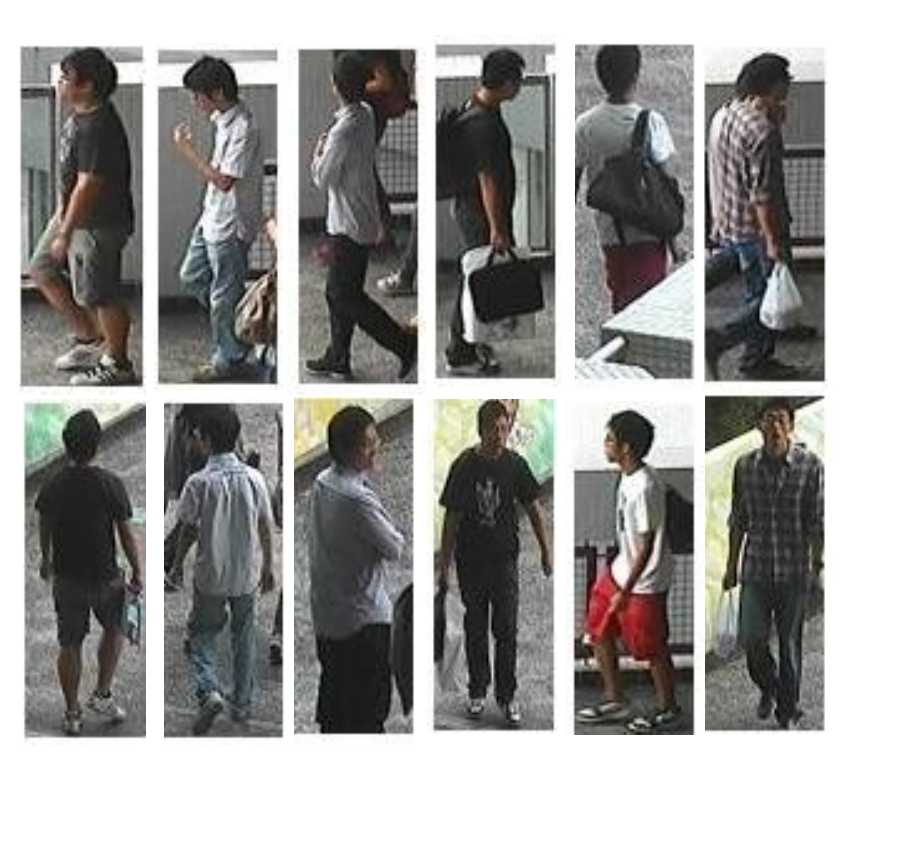}
    \caption{Typical samples of pedestrain images in person re-identification from CUHK03 data set \cite{FPNN}. Each column shows two images of the same individual observed by two different camera views. }\label{fig:example}
\end{figure}

\IEEEPARstart{T}{he}  task of person re-identification (re-id) is to match pedestrian images observed from multiple non-overlapping
camera views with varied visual features \cite{Gheissari2006Person,Roth2014Mahalanobis,MidLevelFilter,Zheng2011Person,Xiong2014Person,Zhao2013Unsupervised,Pedagadi2013Local,LADF,Zhao2013SalMatch}.
 Challenges are presented in the form of compounded variations in visual appearance across different camera views, human poses, illuminations, background clutter, occlusions, relatively low resolution and the different placement of the cameras. Some exmaples are shown in Fig.~\ref{fig:example}.

Person re-identification is essentially to measure the similarity for pairs of pedestrian images in a way that a pair is assigned a high similarity score in case of depicting the same identity and a low score if displaying different identities. This typically involves constructing a robust feature representation and an appropriate similarity measure in order to estimate accurate similarity scores. To this end, many methods focusing on feature representation and distance function learning are designed separately or jointly to deal with person re-id problem. Low-level features such as color and texture can be used for this purpose. Some studies have obtained more distinctive and reliable feature representations,
including symmetry-driven accumulation \cite{Farenzena2010Person}, horizontal partition \cite{LOMOMetric,Zheng2013PAMI}, and salience matching \cite{Zhao2013Unsupervised,Zhao2013SalMatch}. However, it is still difficult to design a type of feature that is discriminative and invariant to severe changes in terms of misalignment across disjoint camera views. Another pipeline of person re-id system is to learn a robust distance or similarity function to deal with complex mathing problem. Many metric leanring algorithms are proposed for this purpose \cite{Pedagadi2013Local,Kedem2012Nonlinear,Kostinger2012Large,Xiong2014Person,LADF,PCCA,Zheng2013PAMI,Hirzer2012Person}.
In practice, most of metric learning methods exhibit a two-stage processing which typically extract hand-crafted features and subsequently learn the metrics. Thus, these approaches often lead to sub-optimal solutions.

Convolutional Neural Networks (CNNs) have proven highly successful at image recognition problems and various surveillance applications including pedestrian detection \cite{DeepContextDetection,DeepDetection}, and tracking \cite{DeepTracking}. However, little progress is witnessed in person re-id, except a few works in \cite{JointRe-id,FPNN,DeepReID}.  By applying CNNs, a joint feature representation and metric learning can be achieved. The FPNN algorithm \cite{FPNN} makes the first attemp to introduce patch matching in CNNs, followed by an improved deep learning framework \cite{JointRe-id} where layers of cross-input neighborhood differences
and patch summary are added. These two methods both evaluate the pair similarity early in the CNN stage, so that it could make use of spatial correspondence of feature maps. In fact, spatial misalignment is very notable in person re-id due to similar appearance or occlusion \cite{ReID_structure}. As a result, a more deep model is demanded to well address this challenge by faithfully capturing non-linear relationship between patches. We aim to improve the state-of-the-art architecture of \cite{JointRe-id} to achieve better accuracy. Specifically, we increase the depth of the network in \cite{JointRe-id} by adding more convolutional layers, which is feasible due to the use of very small (3$\times$3) convolution filters in all layers.

By increasing the depth of AlexNet \cite{Krizhevsky2012Imagenet} using an architecture with very small convolution filters,VGG network \cite{VGG} has shown a significant improvement over the prior-art configurations and generalise well to other databases. Encouraged by these positive results from VGG model, we deepen a state-of-the-art network \cite{JointRe-id} on person re-id task while achieving notable improvement. Our PersonNet consists of ten layers with weights and very small 3$\times$3 receptive fields throughout the whole net.
In training stage, we dynamically sample pairs of images in an online manner where the magnitudes of gradients can very widely for different layers, especially in very deep nets. To this end, we introdue an adaptive root-mean-square (RMS) gradient decent algorithm, RMSProp \cite{rmsprop,GenerateSeq}, which works by dividing the gradient by a running average of its recent magnitude. This adaptive gradient decent algorithm is more suitable to deep layers, and converges much faster than Stochastic Gradient Decent (SGD).
We illustrate the architecture in Section \ref{sec:architecture}.

The main contributions of this paper are three-fold:
\begin{itemize}
\item We present a deep network of increasing depth using an architecture with very small (3$\times$3) convolution filters, which shows a notable improvement on person re-id by pushing the depth to 7-10 weight layers.
\item We employ a different mini-batch gradient decent algorithm in back propagation, RMSProp, which adjusts the magnitudes of the gradients in each mini-batch sampled online. The integration of RMSProp makes our network reach convergence more quickly.
\item Extensive experiments are conducted on benchmark datasets to validate the effectiveness of our architecture.
  We achieve {\em the best reported results} on three popular benchmark datasets.
\end{itemize}

\section{Related Work}\label{sec:related}

Many recent studies on person re-identification attempt to generate robust feature representation which is discriminative and robust for describing a pedestrian's appearance under various changes and conditions
 \cite{Bazzani2012Multiple,Farenzena2010Person,Gheissari2006Person,Gray2008Viewpoint,Wang2007Shape,Zhao2013Unsupervised,MidLevelFilter}.
Bazzani \etal \cite{Bazzani2012Multiple} represent a person by a global mean color histogram and recurrent local patterns through epitomic analysis.
Farenzena \etal \cite{Farenzena2010Person} propose the symmetry-driven accumulation of local features (SDALF) which exploits both symmetry and asymmetry, and represents each part of a person by a weighted color histogram, maximally stable color regions and texture information. Gray and Tao \cite{Gray2008Viewpoint} propose to use AdaBoost to select good features out of a set of color and texture features. Schwartz and Davis propose a discriminative appearance based model using partial least squares, in which multiple visual features: texture, gradient and color features are combined \cite{Schwartz2009Learning}. 
Recently, saliency information has been investigated for person re-id \cite{Zhao2013Unsupervised,Zhao2013SalMatch}, leading to a novel feature representation and improved discriminative power in person re-id.
In \cite{Zhao2013Unsupervised}, a method of (eSDC) is presented to learn salience for persons under deformation.
Moreover, salience matching and patch matching can be integrated into a unified RankSVM framework (SalMatch \cite{Zhao2013SalMatch}).
They also propose mid-level filters (MidLevel) for person re-identification
by exploring the partial area under the ROC curve (pAUC) score \cite{MidLevelFilter}. Lisanti \etal \cite{Giuseppe2015PAMI} leverage  low-level feature descriptors to approximate the appearance variants in order to discriminate individuals by using sparse linear reconstruction model.

Metric learning approaches to person re-id is  to essentially formalize the problem as a supervised metric/distance learning where a projection matrix is sought out so that the projected Mahalanobis-like distance is small when feature vectors represent the same person and large otherwise. Among many metric learning methods, Large Margin Nearest Neighbor Learning (LMNN) \cite{Weinberger2006Distance}, Information Theoretic Metric Learning (ITML) \cite{Davis2007Information}, and Logistic Discriminant Metric Learning (LDM) \cite{Guillaumin2009Isthatyou} are three representative methods. 
By applying these metric learning methods into person re-id, many effective approaches are developed \cite{Pedagadi2013Local,Kedem2012Nonlinear,Kostinger2012Large,Wu2011Optimizing,Xiong2014Person,LADF,PCCA,Zheng2013PAMI,Hirzer2012Person}. 
Mignon \etal \cite{PCCA} proposed Pairwise Pairwise Constrained Component Analysis (PCCA) to learn a projection into a low dimensional space in which the distance between pairs of samples respects the desired constraints, exhibiting good generalization properties in the presence of high dimensional data. Zheng \etal \cite{Zheng2013PAMI} presented a Relative Distance Comparision (RDC) to maximize the likelihood of a pair of true matches having a relatively smaller distance than that of a mismatched pair in a soft discriminant manner.
Koestinger \etal propose the large-scale metric learning from equivalence constraint (KISSME) which considers a log likelihood ratio test of two Gaussian distributions \cite{Kostinger2012Large}.
Li \etal propose the learning of locally adaptive decision functions (LADF), which can be viewed as a joint model of distance metrics and locally adapted thresholding rules \cite{LADF}. The Cross-view Quadratic Discriminant Analysis (XQDA) algorithm learns a discriminant subspace and a distance metric simultaneously, which is able to perform dimension reduction and select the optimal dimensionality.  To make the metric learning more efficient, they further present a  positive semidifinite constrained method to reduce the computation cost and get more robust learned metric \cite{ReID_structure}. In \cite{LOMOMetric}, an efficient feature representation called Local Maximal Occurrence is proposed, followed by a subspace and metric learning method. Last but not least, learning to rank can be employed in person re-id, and approaches include ensembled RankSVM \cite{Prosser2010Person}, Metric Learning to Rank (MLR) \cite{McFee2010Metric} and its application to person re-id \cite{Wu2011Optimizing} and structured metric ensembles \cite{paul2015ensemble}.

Three deep learning based person re-id algorithms have been proposed \cite{DeepReID,JointRe-id,FPNN}. Yi \etal \cite{DeepReID} utilized a Siamese CNN with a symmetry structure  comprising two independent sub-nets, and then employed consine distance as their metric. Li \etal \cite{FPNN} designed a different network, which begins with a single convolution layer with max pooling, followed by a patch-matching layer that multiplies convolutional feature responses from the two inputs at a variety of horizonal offsets. The most similar work to us is JointRe-id \cite{JointRe-id} wher a layer of computing cross-input neighborhood difference features is introduced after two layers of convolution and max pooling.

Our architecture differs substantially from these previous networks. The network is very deep with very small (3$\times$3) convolution filters. This has a significant improvement based on the prior-art configuration by pushing the depth to 7-10 weight layers. Moreover, an adaptive gradient decent algorith, RMSProp is used in our network, which can be immune to initialisation and the instability of gradient in deep nets. Consequently, our network outperforms all previous approaches on the largest Market-1501 data set \cite{Market1501}, CUHK03 \cite{FPNN} and smaller CUHK01 datasets \cite{GenericMetric}.

\section{The architecture}\label{sec:architecture}

\begin{figure*}[hbt]
    \centering
        \includegraphics[width=7in,height=3.5in]{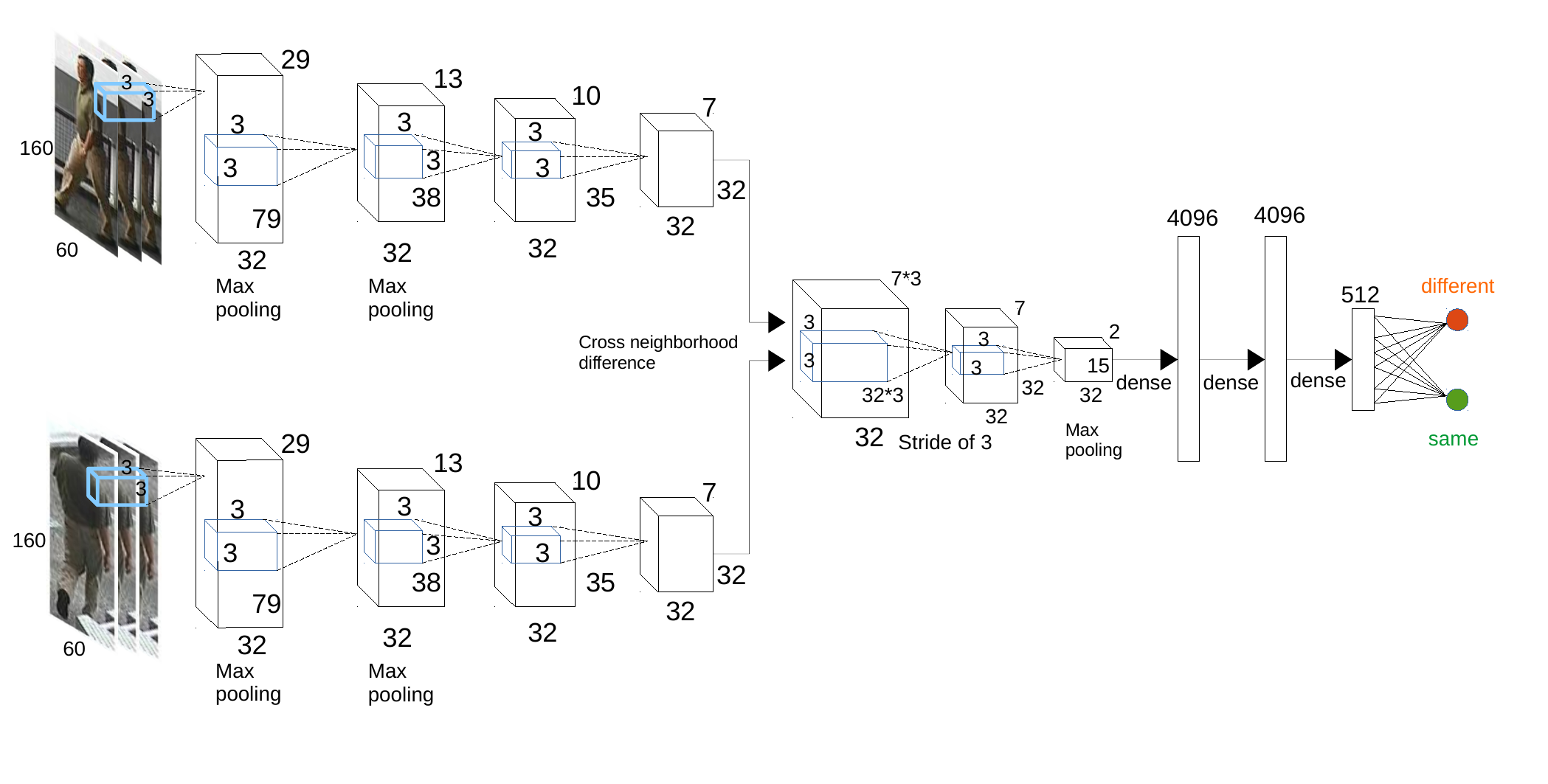}
    \caption{The architecture of PersonNet. The network takes a pair of RGB images as input, which is put through a stack of convolution layers, matching layer, and higher layers computing relationships between them. The configurations of each layer are shown in Table \ref{tab:layer}.}\label{fig:framework}
\end{figure*}

During training, the input to our PersonNet is a pair of fixed-size 160$\times$60 RGB images. The pair of images is passed through a stack of tied convolutional layers, where we use filters with a very small receptive filed: 3$\times$3. The convolution stride is fixed to 1 pixel. Spatial pooling is carried out by three max-pooling layers, which follow some of the convolution layers (not all the convolution layers are followed by max-pooling). Max-pooling is performed over a 2$\times$2 pixel window, with stride 2. After a stack of convolution layers, we have three fully-connceted layers where the first two have 4096 dimension and the third is 512, which is put through softmax to determine the pair is same or different.The overall architecure of the proposed PersonNet is shown in Fig.\ref{fig:framework}.  We show details of these layers in Table \ref{tab:layer}.

\begin{table}[t]
  \centering
  \caption{Layer parameters of PersonNet. The output dimension is given by height$\times$width$\times$width. FS: filter size for convolutions. Layer types: C: convolution, MP: max-pooling, FC: fully-connected.  All convolution and FC layers use hyperbolic tangent as activation function.}  \label{tab:layer}
  {
  \begin{tabular}{c|c|c|c|c}
  \hline
\hline
    Name  & Type  &  Output Dim & FS  & Stride \\
  \hline\hline
   Conv0  & C & $157\times57\times32$  & 3$\times3$ & 1 \\
   Pool0  & MP &  $79\times 29\times 32$ & 2$\times2$ & 2 \\
   Conv1  & C & $76\times 26\times 32$ & 3$\times$3 & 1 \\
   Pool1  & MP & $38\times 13\times 32$ & 2$\times$2 & 2 \\
   Conv2  & C &  $35\times 10\times 32$ &  3$\times$3  & 1 \\
   Conv3  & C & $32\times 7\times 32$ & 3$\times$3 &  1\\
   Difference  & - & $32\times 7\times 32$ & 3$\times$3 &  3\\
   Conv4  & C & $32\times 7\times 32$ & 3$\times$3 &  1\\
   Pool4  & MP & $15\times 2\times 32$ & 2$\times$2 &  2\\
\hline
   FC1   & FC  & -  & 4096 &  - \\
   FC2   & FC  & -  & 4096  & -  \\
   FC3   & FC  & - &  512 & -  \\
  \hline
  \end{tabular}
  }
\end{table}

\subsection{Convolution and max pooling}

The first two layers are concolutional and max-pooling layers. Given two pedestrain images $\bI$ and $\bJ$ observed by two different camera views with three color channels and sized $160\times 60$, the convolutional layer outputs local features extracted by filter paired. The filters $(\bW,\bV)$ applied to two camera views are shared. Given the input $\bI$ and $\bJ$, consisting of $C$ channels of height $H$ and width $W$, if we use $K$ filters and each filter is in size of $m\times m \times C$, the output consits of a set of $C'$ channels of height $H'$ and width $W'$. We define the filter functions as $f, h: \mathbb{R}^{H\times W \times C} \rightarrow \mathbb{R}^{H' \times W' \times C'}$
\begin{align}
f_{ij}^k = \sigma((\bW_k * \bI)_{ij} + b_k^I)\\\nonumber
h_{ij}^k = \sigma((\bV_k * \bI)_{ij} + b_k^J),\nonumber
\end{align}
where $k\in \mathbb{R}^{K\times K\times C}$. Rather than using relatively large receptive fields in the first two covolutional layers (e.g., 5$\times$5 in \cite{JointRe-id,FPNN}), , we use small receptive fileds with 3$\times$3 throughout the whole net to convolute with the input at every pixel with stride of 1. Apparently, a stack of two 3$\times$3 convolution layers (without spatial pooling between them) amounts to working as a receptive field of 5$\times$5. By doing this, more non-linear activation functions are embeded which can make the decision function more discriminative \cite{VGG}.

Activation function can increase the nonlinear properties of the decision function and of the overall network without affecting the receptive fields of the convolutional layer. Insead of using ReLU, $\sigma(\bx)=\max(0,\bx)$, as the  activation function in deep network, we choose the nonlinear activation function $\sigma(\bx)$ to be hyperbolic tangent function, $\sigma(\bx)=tanh(\bx)=\frac{e^{x} - e^{-x}}{e^{x} + e^{-x}}$, which can rescale the linear output in the range [-1, 1].  Scaling the activation function to be $\sigma(\bx)=tanh(\frac{3\bx}{2})$ is able to ensure the training is spread uniformly over each layer particularly helpful in networks with very deep layering.

The max-pooling layer is used to reduce the dimensionality of the output and variance in deformable objects to ensure that the same result will be obtained even when image features undergo slight translations. The max-pooling operation is applied on every pixel around its  neighborhood.

\subsection{Modeling neighborhood patch matching}\label{ssec:neighbor}

The patch matching layer is to computes the differences in filter responses of local pathes across two views. Since we have $f_I, g_J \in \mathbb{R}^{32\times7}$, the difference around the neighborhood of each feature location yields to a set of feature maps $K_i\in \mathbb{R}^{32\times7\times3\times3}$ ($i=1,\ldots,32$), where $3\times3$ is the window size of neighborhood around a feature value. In other words, $K_i$ indicates a $32\times7$ grid of $3\times3$ blocks, $K_i(x,y)\in \mathbb{R}^{3\times3}$ where $1\leqslant x \leqslant 7$ and $1\leqslant y\leqslant 32$. Following \cite{JointRe-id}, we have
\begin{align}
K_i(x,y)=f_I(x,y)\mathbb{I}(3,3)-\mathcal{N}[h_J(x,y)]
\end{align}
where $\mathbb{I}\in \mathbb{R}^{3\times3}$ is a $3\times3$ indicator matrix  with all elements being 1s, and $\mathcal{N}[h_J(x,y)]\in \mathbb{R}^{3\times3}$ is a 3$\times$3 neighborhood of $h_J$ centered at $(x,y)$. Here we use a small neighborhood of size $3\times3$ to model the displacement of body parts caused by pose and viewpoint variations \cite{FPNN}. Our architecure can avoid symmetric operation on computing $K_i$ because we use online sampling to generate pairs of images.

\begin{figure}[hbt]
    \centering
        \includegraphics[width=3.5in,height=1.5in]{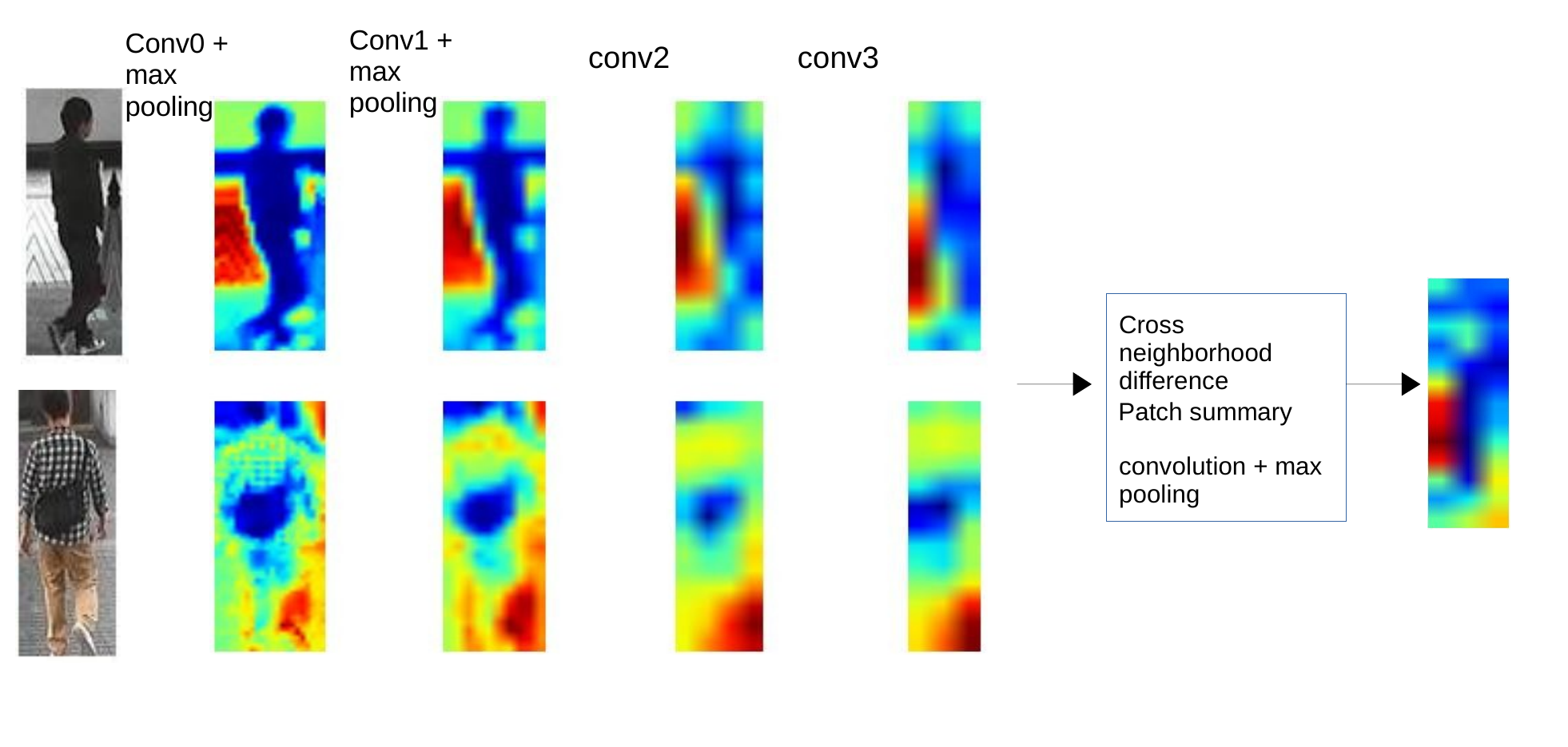}\\
	(a) A pair of negative images\\
	\includegraphics[width=3.5in,height=1.5in]{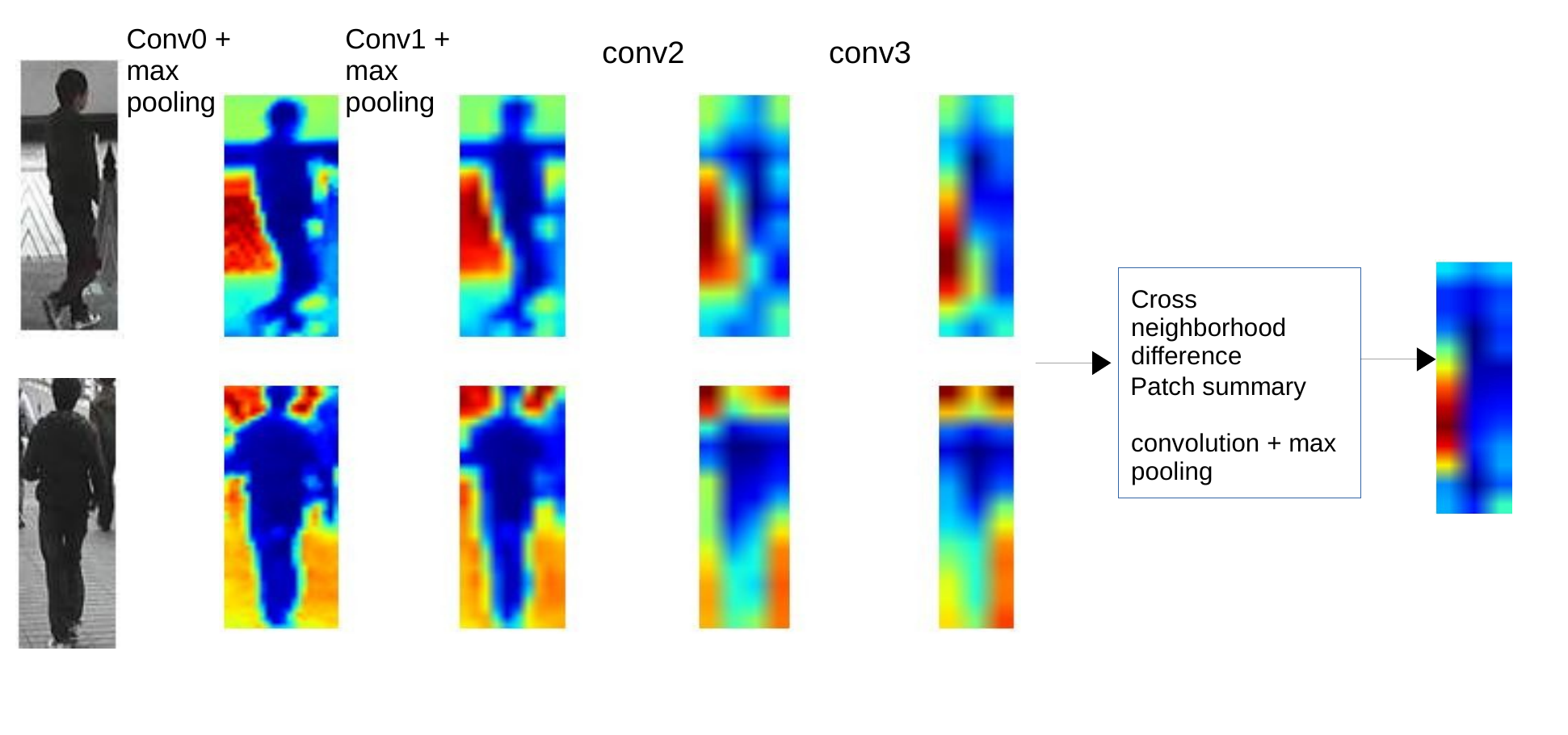}\\
	(b) A pair of positive images
    \caption{Feature responses of each layer learned by our network. See Section \ref{sec:architecture} for details. }\label{fig:neg_pos}
\end{figure}

The visualization of feature responses at each layer of the network are shown in Fig.\ref{fig:neg_pos}. We can see that after Conv0, the features responds to bright regions of the images. After a few convolutions and max-pooling, higher responses are given to body as a whole. In this process, part-based CNNs maybe beneficial to further improve the accuracy of recognition since human body parst can be very different across camera views and matching different parts are helpful in matching. Recall from \ref{ssec:neighbor} and \cite{JointRe-id}, the neighborhood layer is to compute the difference of corresponding feature maps across two views in a small range. This can robustly match some patches that undergo variations in viewpoints, and poses. For a negative pair, a neighborhood difference layer can highlight some local pathes that are visually different, as shown in Fig.\ref{fig:neg_pos} (a). By contrast, for a positive pair,  the difference map is expected to be close to zero and nonzero values should be small and uniformly distribute across the map, as shown in Fig.\ref{fig:neg_pos} (b). The difference layer is followed by another patch summary layer that extract these difference maps into a holistic representation of of the differences in each $3\times3$ block. Then, we use another convolution layer with max pooling to learn spatial relationships across neighborhood differences. The network ends up with three fully connected layers with softmax output.

\section{Training strategies}\label{sec:training}

We use the hyperbolic tangent function as the non-linear activation function in our models.
Trainning data are divided into mini-batches. Note that we employ RMSProp \cite{rmsprop} instead of commonly used stochastic gradient decent (SGD) as a means of updating gradients for parameters.

\subsection{RMSProp}
RMSProp works by dividing the gradient by a running average of its recent magnitude.  The main difference between SGD and RMSProp stems from the way of making use of gradient. The idea behind SGD is that when the learning rate is small, it averages the gradients over successive mini-batches. However, the magnitude of the gradient can be very different for different weights and can change during learning. Hence, it is impratical to choose a single learning rate. To work efficiently with mini-batches, RMSProp is developed to use the gradient but divide by a different number for each mini-batch and the number divided by is similar for adjacent mini-batches.
Thus, RMSProp keeps the a moving average of the squared gradient for each weight:
\begin{equation}\scriptsize
\begin{aligned}
& MeanSquare(\bw,t)=0.9*MeanSquare(\bw,t-1) + 0.1* \left(\frac{\partial E}{\partial \bw^{(t)}}\right)^2\\\nonumber
& \partial \bw^{(t)}=\epsilon \frac{\partial E}{\partial \bw^{(t)}}/\left( MeanSquare(\bw,t)^{\frac{1}{2}}+ \mu \right)
\end{aligned}
\end{equation}
where $\bw$ is the weight parameter, $t$ is the time step, $\epsilon$ is the learning rate, $\mu$ is a smoothing value for numerical convention, and $E$ denotes the error surface. A recent study \cite{rmsprop} has shown that dividing the gradient by $\left(MeanSquare(\bw,t)\right)^{\frac{1}{2}}$  makes the learning work much better.  The introduction of RMSProp is beneficial to our architecture, which performs more robustly than SGD.

\subsection{Data augmentation and data balancing}

In the training set, the matched (positive) pairs are several orders fewer than non-matched (negative) pairs, which can lead to data imbalance and overfitting. To circumvent this issue, we augment data set by performing 2D translation on each pedestrian image. Specifically, following \cite{FPNN,JointRe-id} for an original image of size $H\times W$, five images of the same size are randomly sampled around the original image center, with translation drawn from a uniform distribution in the range $[-0.05H,0.05H]\times [-0.05W,0.05W]$. For CUHK01 dataset, we also horizonally relfect each image.

To achieve data balancing, we online sample the same number of negative and positive pairs instead of generating the proportion of negative pairs against positives in a fixed manner. For example, a common way as conducted in \cite{FPNN,JointRe-id} is to generate the same size of negatives and positives, then gradually increase the number of negative samples up to the ratio of 5:1. Such operation is unable to learn a robust and reliable network that tolerate the varied data distributions in each mini-batch. Nonetheless, our online sampling strategy can well address  the aspect of data balance.

\section{Experimental results}\label{sec:exp}

We implemented our network using Theano deep learning framework \cite{Theano}.
The training of the network converges in roughly 20-22 hours in NVIDIA GeForce GTX 980 GPU. The training is carried out by optimising the softmax objective using online sampling of each input pair of images with RMSProp gradient decent. The mini-batch size was set to 2. The training was regularised by $L_2$ penalty and dropout \cite{Krizhevsky2012Imagenet} regularisation for the first two fully-connected layers (dropout ratio set to 0.5) in order to alleviate over-fitting. The learing rate was initially set to 0.05, and then decreased by a factor of 10 when the validation set accuracy stopped improving. In general, the learning was stopped within 100K iterations. We conjecture that inspite of the larger number of parameters and the greater depth of our nets compared with FPNN and JointRe-id, the nets required less iterations to converge due to (a) the implicit regularisation induced by greater depth and smaller convolution sizes; (b) adaptive gradient decent algorithm of RMSProp.

\begin{figure}[hbt]
    \centering
        \includegraphics[width=2.5in,height=2in]{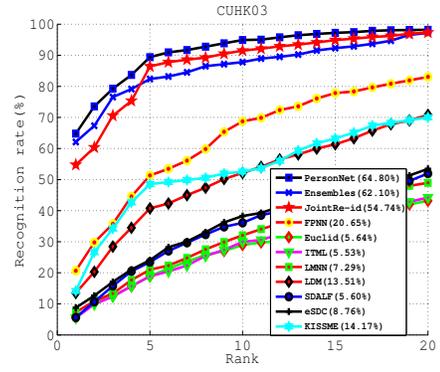}\\
	(a)  CUHK03 data set\\
	\includegraphics[width=2.5in,height=2in]{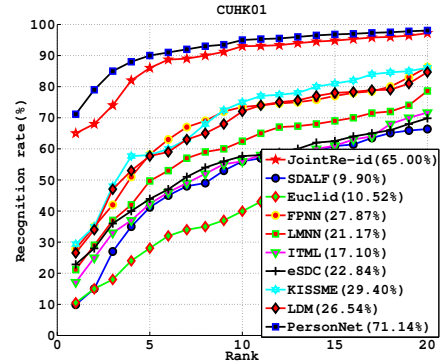}\\
	(b) CUHK01 data set
    \caption{Performance comparison with state-of-the-art approaches using CMC curves on CHK03 and CUHK01 datasets.}\label{fig:cmc_results}
\end{figure}

\subsection{Experimental settings}

\paragraph{Datasets.} We perform experiments on three benchmarks: the CUHK03 dataset \cite{FPNN}, the CUHK01 dataset \cite{GenericMetric} and the Market-1501 dataset \cite{Market1501}.

\paragraph{Evaluation protocol} We adopt the widely used single-shot modality in our experiment to allow extensive comparison. Each probe image is matched against the gallery set, and the rank of the true match is obtained. The rank-$k$ recognition rate is the expectation of the matches at rank $k$, and the cumulative values of the recognition rate at all ranks are recorded as the one-trial Cumulative Matching Characteristice (CMC) results \cite{paul2015ensemble}. This evaluation is performed ten times, and the average CMC results are reported.

\paragraph{Competitors} We compare our model with the following state-of-the-art approaches: SDALF \cite{Farenzena2010Person}, ELF \cite{Gray2008Viewpoint}, LMNN \cite{Hirzer2012Person}, ITML \cite{Davis2007Information}, LDM \cite{Guillaumin2009Isthatyou}, eSDC \cite{Zhao2013Unsupervised}, SalMatch \cite{Zhao2013SalMatch}, Generic Metric \cite{GenericMetric}, Mid-Level Filter (MLF) \cite{MidLevelFilter}, eBiCov \cite{eBiCov}, PCCA \cite{PCCA}, LADF \cite{LADF}, kLFDA \cite{Xiong2014Person}, rPCCA \cite{Xiong2014Person}, RDC \cite{Zheng2013PAMI}, RankSVM \cite{RankSVM}, Metric Ensembles (Ensembles) \cite{paul2015ensemble}, KISSME \cite{Kostinger2012Large}, JointRe-id \cite{JointRe-id}, FPNN \cite{FPNN}.

\subsection{Experiments on CUHK03 data set}

\begin{table}[t]
  \centering
  \caption{Rank-1, Rank-5, and Rank-10 recognition rate of various methods over CUHK03 dataset. }  \label{tab:cmc_cuhk03}
  {
  \begin{tabular}{c|c|c|c|c}
  \hline
\hline
    Method  & $r=1$  &  $r=5$ & $r=10$  & $r=20$ \\
  \hline\hline
   Ensembles \cite{paul2015ensemble}  & $62.10$ &  87.81 & 92.30 & 97.20 \\
   JointRe-id \cite{JointRe-id}  & $54.74$ & 86.42 & 91.50 & 97.31 \\
   FPNN \cite{FPNN} & $20.65$ & 51.32 & 68.74 & 83.06\\
   Euclid  & $5.64$ & 18.93 &28.96 & 43.17 \\
   ITML \cite{Davis2007Information} & $5.53$ & 18.89 &29.96 & 44.20 \\
   LMNN \cite{Hirzer2012Person} & $7.29$ & 21.00 & 32.06& 48.94 \\
   LDM \cite{Guillaumin2009Isthatyou} & $13.51$ & 40.73 &52.13 & 70.81 \\
   SDALF \cite{Farenzena2010Person} & $5.60$ & 23.45 &36.09 & 51.96 \\
   eSDC \cite{Zhao2013Unsupervised} & $8.76$ & 24.07 &38.28 & 53.44 \\
   KISSME \cite{Kostinger2012Large} & $14.17$ & 48.54 &52.57 & 70.03 \\
\hline
   PersonNet   &  $\mathbf{64.80}$ &  $\mathbf{89.40}$ & $\mathbf{94.92}$ &  $\mathbf{98.20}$ \\
  \hline
  \end{tabular}
  }
\end{table}

The CUHK03 dataset includes 13,164 images of 1360 pedestrians. The whole dataset is captured with six surveillance camera. Each identity is observed by two disjoint camera views, yielding an average 4.8 images in each view. This dataset provides both manually labeled pedestrian bounding boxes and bounding boxes automatically obtained by running a pedestrian detector \cite{DetectionPAMI}. In our experiment, we report results on labeled data set. As can be seen from Fig.\ref{fig:cmc_results} (a) and Table \ref{tab:cmc_cuhk03}, our very deep PersonNet outperform the previous methods, which particularly improves from 62.1\% (Ensemble \cite{paul2015ensemble}) to 64.8\%.

\subsection{Experiments on CUHK01 data set}

\begin{table}[t]
  \centering
  \caption{Rank-1, Rank-5, and Rank-10 recognition rate of various methods over CUHK01 dataset. }  \label{tab:cmc_cuhk01}
  {
  \begin{tabular}{c|c|c|c|c}
  \hline
\hline
    Method  & $r=1$  & $ r=5 $& $r=10 $ & $r=20$ \\
  \hline\hline
   JointRe-id \cite{JointRe-id}  & $65.00$ & $88.70$ &93.12 & $97.20$ \\
   SDALF \cite{Farenzena2010Person} & 9.90 & 41.21 & 56.00 & 66.37 \\
   Euclid  & 10.52 & 28.07 &3 9.94 & 55.07 \\
   FPNN \cite{FPNN} & 27.87 & 58.20 & 73.46 & 86.31 \\
   LMNN \cite{Hirzer2012Person} & 21.17 & 49.67 & 62.47 & 78.62 \\
   ITML \cite{Davis2007Information} & 17.10 & 42.31 & 55.07 & 71.65 \\
   eSDC \cite{Zhao2013Unsupervised} & 22.84 & 43.89 & 57.67 & 69.84 \\
   KISSME \cite{Kostinger2012Large}  & 29.40 & 57.67 &72.43\% & 86.07 \\
   LDM\cite{Guillaumin2009Isthatyou}  & 26.45 & 57.69 &72.04\% & 84.69 \\
\hline
   PersonNet   &  $\mathbf{71.14}$ &  $\mathbf{90.07}$ & $\mathbf{95.00}$ &  $\mathbf{98.06}$ \\
  \hline
  \end{tabular}
  }
\end{table}

The CUHK01 data set has 971 identities with 2 images per person in each view. We report results on the setting where 100 identities are used for testing, and the remaining 871 identities used for training, in accordance with FPNN \cite{FPNN}. Fig. \ref{fig:cmc_results} (b) and Table \ref{tab:cmc_cuhk01} compare the performance of our model with previous methods.
Our approach is superior to the state of the art by a large margin with a rank-1 recognition rate of 71.14\% against 65\% by the next best method.

\subsection{Experiments on Market-1501 data set}

\begin{table}[t]
  \centering
  \caption{Rank-1 and  mAP of various methods over Market-1501 dataset. }  \label{tab:cmc_market}
  {
  \begin{tabular}{c|c|c}
  \hline
\hline
    Method  & $ r=1$  &  mAP \\
  \hline\hline
   SDALF \cite{Farenzena2010Person} & 20.53 & 8.20\\
   eSDC \cite{Zhao2013Unsupervised} & 33.54 & 13.54\\
   Zheng \etal \cite{Market1501} & 34.40 &  14.09\\
\hline
   PersonNet   & $\mathbf{37.21}$  &  $\mathbf{18.57}$ \\
  \hline
  \end{tabular}
  }
\end{table}

The Market-1501 dataset contains 32,643 fully annoated boxes of 1501 pedestrians, making it the largest person re-id dataset to date. Each identity is captured by at most six cameras and boxes of person are obtained by running a state-of-the-art detector, the Deformable Part Model (DPM) \cite{MarketDetector}.
As conducted in, the dataset is randomly divided into training and testing sets, containing 750 and 751  identities, respectively.

We compare our model with state-of-the-art methods in Table \ref{tab:cmc_market}. The results are reported on single-shot and single-query. We can see that our deep network outperforms these methods notably on Market-1501 dataset.

\subsection{Convergence study}

\begin{figure*}[hbt]
    \centering
        \includegraphics[width=2.2in,height=1.8in]{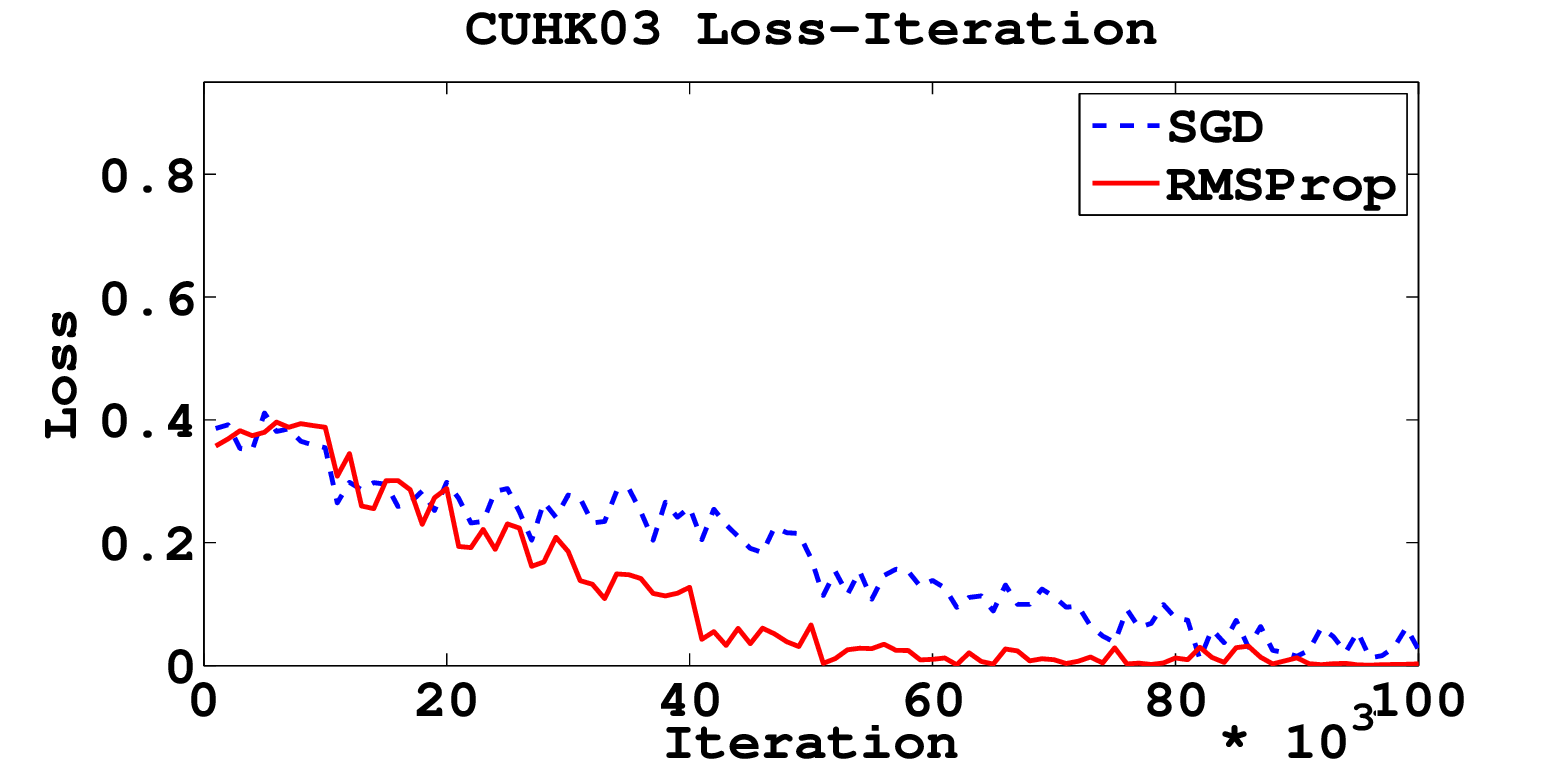}
        \includegraphics[width=2.2in,height=1.8in]{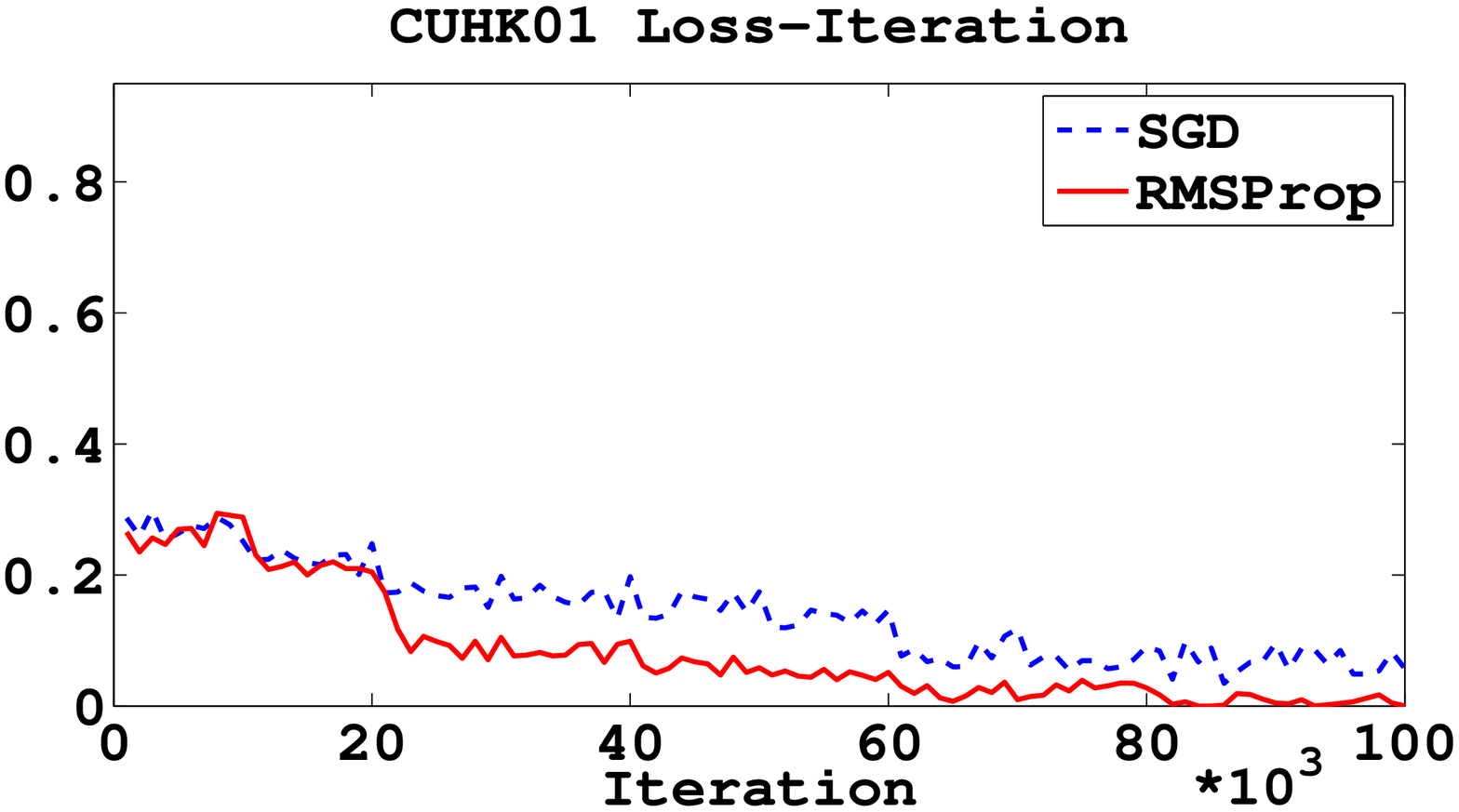}
	\includegraphics[width=2.2in,height=1.8in]{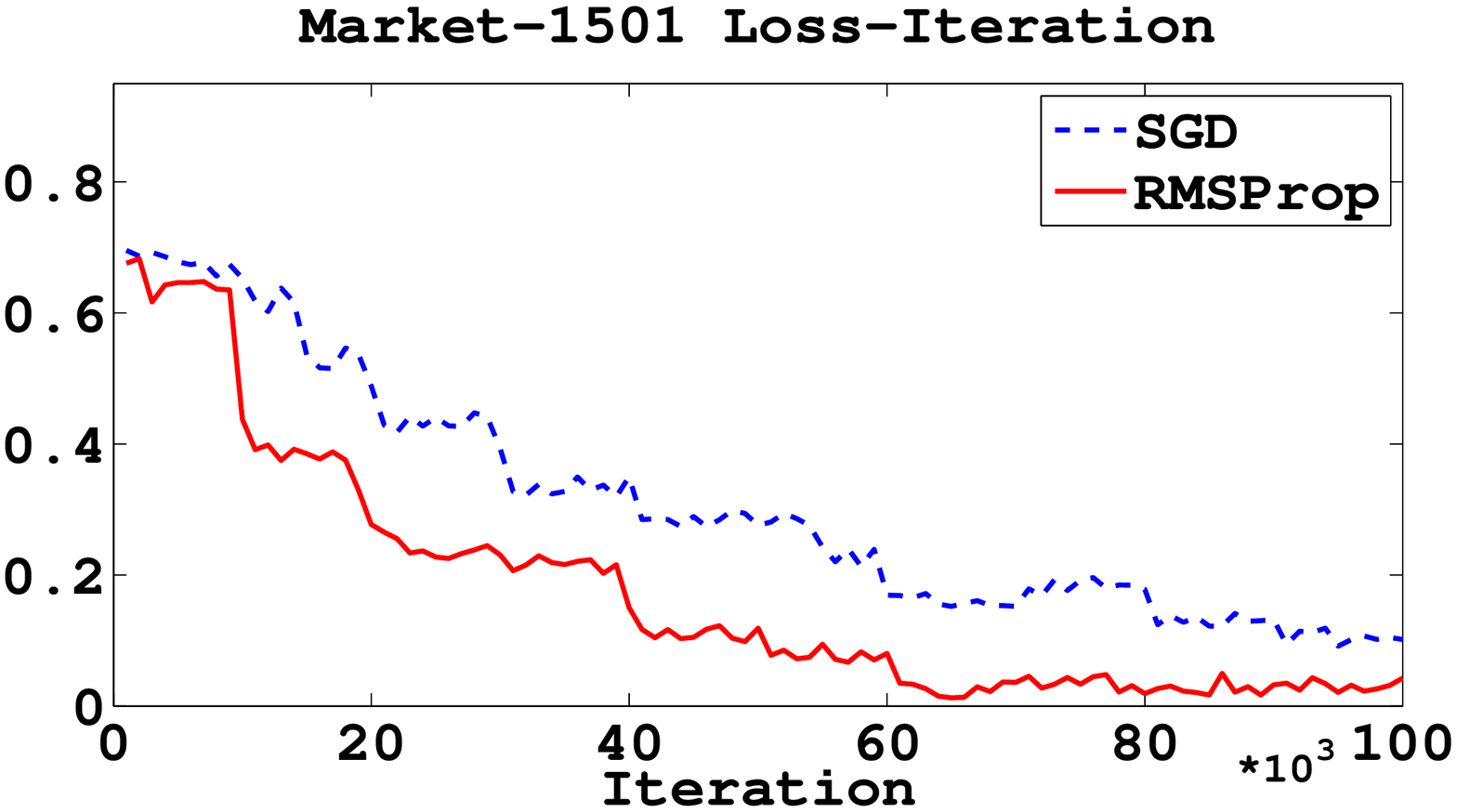}\\
    \caption{Study on convergence speed of SGD and RMSProp.}\label{fig:convergence}
\end{figure*}

In this section, we study the convergence speed of SGD and RMSProp and report empirical results in Fig. \ref{fig:convergence}. It can be seen that RMSProp is more stabely and relatively faster to be converged than SGD. This is mainly because SGD as itself  is solely depending on the given batch of instances of the present iteration. Therefore, it tends to have unstable update steps per iteration and convergence takes more time or even get stuck into local minima. By contrast, RMSProp keeps running average of its recent gradient magnitudes and divides the next gradient by this average so that loosely gradient values are normalized. Consequently, RMSProp works better on gradient updates in steps of different batches.

\section{Conclusion}\label{sec:con}

In this work, we evaluated very deep convolutional networks (up to 10 weight layers) for person re-identification. It was demonstrated that the representation depth is beneficial to the recognition accuracy in person matching, and state-of-the-art performance on person re-id datasets including CUHK03, CUHK01, and Market-1501 datasets can be achieved using an effective matching based architecture \cite{FPNN,JointRe-id} with notably increased depth. Our experimental results justify the importance of depth in person identity matching.

\ifCLASSOPTIONcaptionsoff
  \newpage
\fi

\bibliographystyle{IEEEtran}\small
\bibliography{CSRef}

\end{document}